\documentclass[12pt,fleqn]{tsp}

\usepackage{amsmath,graphicx,hyperref}
\usepackage{amssymb}
\usepackage{textcomp}
\usepackage{adjustbox}
\usepackage{xcolor}
\usepackage{booktabs}
\usepackage{multirow}
\usepackage[inkscapelatex=false]{svg}

\newcommand{\tbl}[1]{\textcolor{black}{#1}}
\newcommand{\tred}[1]{\textcolor{black}{#1}}
\newcommand{\bbl}[1]{\textcolor{black}{#1}}
\usepackage{tikz}
\usetikzlibrary{arrows.meta, positioning, fit, calc}

\techscience{vol.}{no.}{2024}{0}
\license{\includegraphics[width=16cm,height=1.5cm]{license}}
\thehead{Journal not specified}%
\newtheoremstyle{note}
{3pt}
{3pt}
{\normalfont}
{}
{\bfseries}
{.}
{.5em}
{}
\graphicspath{{./image/}}
\runningtitle{Manuscript Preparation for TSP} 

\title{Context Patch Fusion With Class Token Enhancement for Weakly Supervised Semantic Segmentation}

\author{Yiyang Fu{\thanks{School of Cyber Science and Engineering, Wuxi University, Wuxi 214105, China.}, Hui Li$^\ast{}$\thanks{School of Informatics, Xiamen University, Xiamen 361005, China}}
and Wangyu Wu$^\ast{}$\thanks{School of Computer Science, University of Liverpool, Liverpool L69 7ZX, UK 
\vskip 0.15cm
{$^\ast{}$Corresponding Author: Hui Li (huilinlp@xmu.edu.cn), Wangyu Wu (Wangyu.wu@liverpool.ac.uk) }\\
  \vskip 0.15cm 
  \vskip 0.15cm
 {Received: Month Day, Year; Accepted: Month Day, Year}}
}

\begin{document}
\maketitle

\begin{mdframed}[backgroundcolor=gray!8,linewidth=0pt,nobreak=true,innerleftmargin=0.2cm,innerrightmargin=0.2cm,leftmargin=-2cm,rightmargin=-2cm]
\begin{abstract}
\addbottompattern

Weakly Supervised Semantic Segmentation (WSSS), which relies only on image-level labels, has attracted significant attention for its cost-effectiveness and scalability. Existing methods mainly enhance inter-class distinctions and employ data augmentation to mitigate semantic ambiguity and reduce spurious activations. However, they often neglect the complex contextual dependencies among image patches, resulting in incomplete local representations and limited segmentation accuracy. To address these issues, we propose the Context Patch Fusion with Class Token Enhancement (CPF-CTE) framework, which exploits contextual relations among patches to enrich feature representations and improve segmentation. At its core, the Contextual-Fusion Bidirectional Long Short-Term Memory (CF-BiLSTM) module captures spatial dependencies between patches and enables bidirectional information flow, yielding a more comprehensive understanding of spatial correlations. This strengthens feature learning and segmentation robustness. Moreover, we introduce learnable class tokens that dynamically encode and refine class-specific semantics, enhancing discriminative capability. By effectively integrating spatial and semantic cues, CPF-CTE produces richer and more accurate representations of image content. Extensive experiments on PASCAL VOC 2012 and MS COCO 2014 validate that CPF-CTE consistently surpasses prior WSSS methods.
\end{abstract}

\end{mdframed}

\vspace{12pt}

\section{Introduction}
\label{sec:intro}

Semantic segmentation is a fundamental task in computer vision, underpinning numerous applications such as autonomous driving, medical image analysis, and remote sensing~\cite{tang2023duat,wang2022stepwise,xiong2024sam2}. Among various paradigms, \textit{Weakly Supervised Semantic Segmentation} (WSSS) has gained increasing attention due to its cost-effectiveness and scalability. Unlike fully supervised approaches that rely on dense pixel-level annotations, WSSS leverages weak labels such as image-level tags~\cite{lee2021anti}, scribbles~\cite{lin2016scribblesup}, or bounding boxes~\cite{lee2021bbam} as alternative supervision. This substantially reduces annotation costs while maintaining competitive segmentation quality.

Early WSSS methods primarily depend on \textit{Class Activation Maps} (CAMs)~\cite{kolesnikov2016seed} to generate pseudo labels from image-level supervision. Despite advances through refined CAM expansion strategies and multi-stage training pipelines~\cite{araslanov2020single,ahn2019weakly,lee2021anti}, these approaches still suffer from two persistent challenges: incomplete object localization and limited segmentation accuracy. The core limitation lies in CAM's tendency to highlight only the most discriminative regions, resulting in fragmented object representations.

Recently, the \textit{Vision Transformer} (ViT)~\cite{dosovitskiy2020image} has revolutionized large-scale image understanding by effectively modeling long-range dependencies across image regions. Leveraging ViT for WSSS has shown promising potential~\cite{ru2023token,xu2024mctformer+}, as transformer-based architectures can capture both local and global contextual relationships. For instance, TOCO~\cite{ru2023token} mitigates the over-smoothing effect in ViT representations, while MCTformer+~\cite{xu2024mctformer+} employ multiple class tokens to enhance class-to-patch attention for improved class-specific localization. Meanwhile, data augmentation strategies~\cite{shorten2019survey} have also proven effective in improving WSSS robustness~\cite{wu2024top,li2023ddaug,wu2023image} by diversifying training data and reducing semantic ambiguity. However, despite these advancements, existing methods still struggle to fully capture the intricate contextual dependencies among image patches, leading to incomplete local representations and suboptimal segmentation performance.

\tred{Although several works explore sequential or recurrent structures to enhance patch dependency modeling, these methods typically operate at early or intermediate stages of the network and do not explicitly resolve the spatial discontinuity caused by ViT patchification.} 
\tred{ \bbl{Unlike previous transformer--RNN fusion approaches~\cite{wensel2023vit,yang2022recurring}, our CF-BiLSTM is explicitly motivated by the patch discontinuity of ViT representations}, and is designed as a post-hoc spatial continuity restoration module acting on globally contextualized features. This design objective has not been explored in prior WSSS literature.}

\tred{Moreover, while multi-class token attention mechanisms (e.g., MCTformer+, AVKT) enrich class--patch interactions inside transformer layers, their early fusion strategy often leads to class competition and entangled feature aggregation.} 
\tred{Furthermore, our class tokens operate at a higher semantic level \emph{after} ViT encoding rather than within early attention blocks, enabling cleaner, non-competing class-specific refinement. This post-hoc semantic enhancement differs fundamentally from the early multi-token attention schemes used in MCTformer+ and AVKT.}

\begin{figure}[t]
\centering
\includegraphics[width=1.0\linewidth]{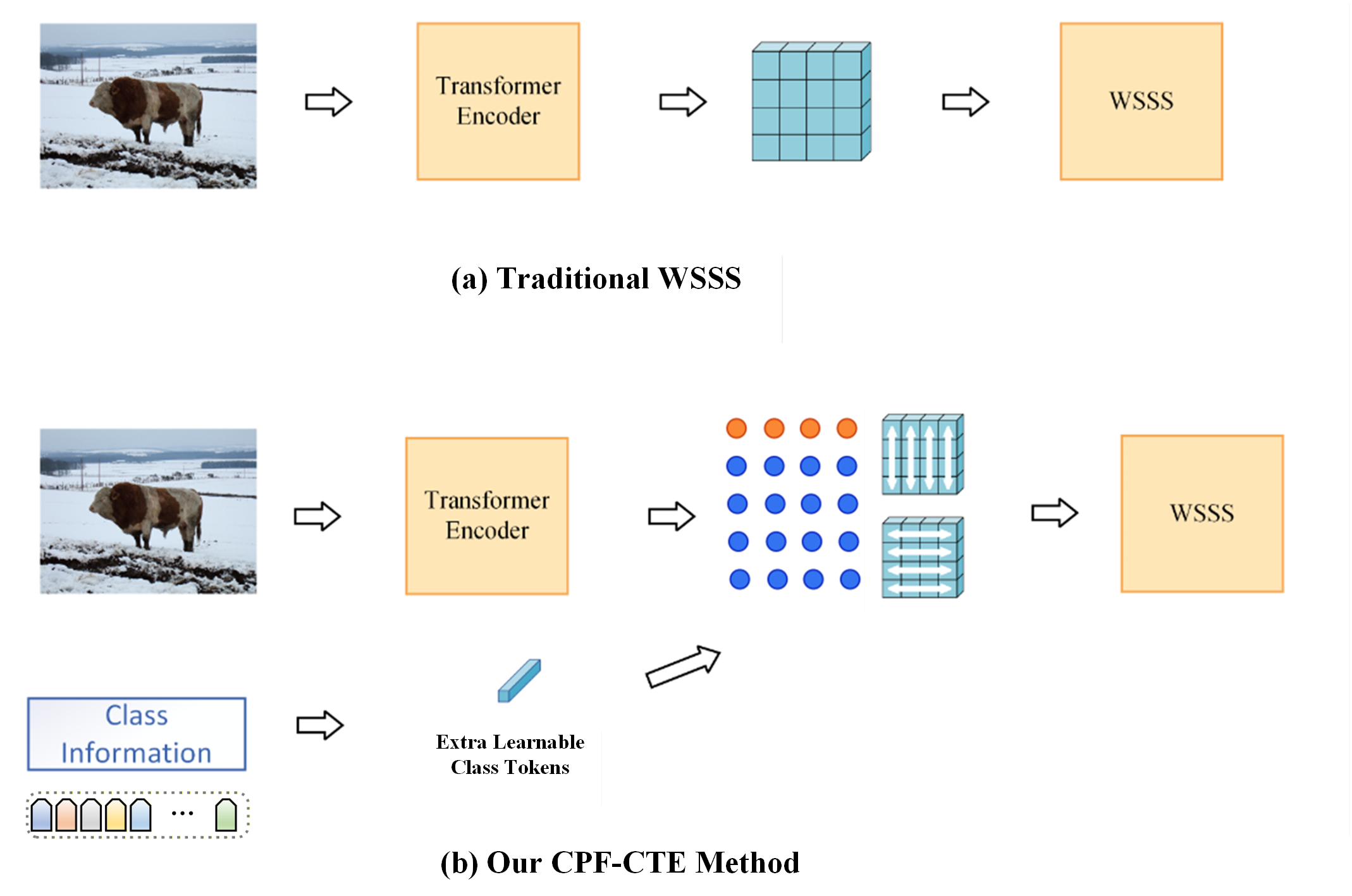}
\caption{\bbl{Comparison between traditional WSSS pipeline and the proposed CPF-CTE framework. (a) Traditional WSSS relies solely on ViT patch embeddings, which often suffer from fragmented and class-ambiguous activations. (b) Our CPF-CTE explicitly incorporates image-level class semantics through learnable class tokens and enhances spatial continuity using the Context Fusion module, leading to more coherent and discriminative feature representations for pseudo-label generation.}}
\label{fig:main}
\vspace{-0.3cm}
\end{figure}

To address these limitations, we propose a novel \textbf{Context Patch Fusion with Class Token Enhancement (CPF-CTE)} framework that fully exploits spatial and semantic relationships among image patches. At its core, the proposed \textit{Contextual-Fusion Bidirectional LSTM (CF-BiLSTM)} module captures bidirectional spatial dependencies between patches, enabling effective inter-patch information flow and contextual reasoning. \tred{In addition, learnable class tokens are introduced \emph{after} ViT encoding to dynamically refine class-specific semantics, avoiding early-stage interference and yielding cleaner semantic conditioning.} \tred{Unlike early class-token injection strategies, which may 
dominate the attention process and introduce class competition during feature 
learning, our post-hoc design intentionally prevents such early semantic 
interference. Since the ViT backbone already provides globally contextualized 
patch embeddings, the class tokens serve as dedicated semantic refiners rather 
than participating in the full-layer attention, enabling cleaner and more 
disentangled class-specific enhancement.}

Comprehensive experiments on the PASCAL VOC 2012~\cite{everingham2010pascal} and MS COCO 2014~\cite{lin2014microsoft} benchmarks demonstrate that CPF-CTE consistently outperforms state-of-the-art (SOTA) WSSS methods. Extensive ablation studies further validate the effectiveness of each component. As summarized in Fig.~\ref{fig:main}, our main contributions are as follows:

\begin{itemize}
    \item We design a \textbf{Contextual-Fusion Bidirectional LSTM (CF-BiLSTM)} module that explicitly models spatial dependencies among image patches, significantly improving inter-patch information exchange and contextual understanding.
    
    \item We introduce \textbf{learnable class tokens} that dynamically encode class-specific semantics through post-hoc refinement, enabling more precise and discriminative patch representations.
    
    \item We develop a \textbf{ViT-based CPF-CTE framework} that jointly leverages spatial context fusion and semantic enhancement, achieving superior performance over existing SOTA WSSS methods on PASCAL VOC 2012~\cite{everingham2010pascal} and MS COCO 2014~\cite{lin2014microsoft}.
\end{itemize}

\section{Related Work}
\label{sec:Relwork}

\subsection{Weakly Supervised Semantic Segmentation}
Most existing Weakly Supervised Semantic Segmentation (WSSS) methods typically follow a three-stage pipeline: First, an initial classification model is trained using image-level labels to generate Class Activation Maps (CAMs)~\cite{kolesnikov2016seed} for training images. Subsequently, these coarse CAMs are iteratively refined through various regularization techniques to produce enhanced pseudo-labels. Finally, a fully supervised semantic segmentation model is trained under the supervision of these refined pseudo-labels. A common drawback of CAMs is that they usually only activate the most discriminative regions of objects. To address this limitation, many works have focused on improving the quality of CAMs to achieve accurate semantic segmentation. Some existing WSSS methods obtain high-quality CAMs through post-processing techniques, such as dense Conditional Random Fields (denseCRF)~\cite{chen2014semantic}, AffinityNet~\cite{ahn2018learning}, or AdvCAM~\cite{lee2021anti}. Additionally, some approaches enhance WSSS performance through data augmentation~\cite{wu2023image} outside the model architecture. However, these refinement strategies are susceptible to noise and imprecise activations and heavily rely on the initial quality of CAMs. ~\cite{yao2021non,lee2021railroad} leverage auxiliary saliency maps to reduce background interference and accurately locate non-salient object parts. Furthermore, ~\cite{chang2020weakly} propose a simple yet effective method to refine CAMs by integrating an unsupervised sub-category identification task. The CAM mechanism generates class-specific localization maps by leveraging pixel-wise associations between class-related weights and image features. However, the limitations of CAMs stem from insufficient class representations. Recent methods aim to address this: IS-CAM~\cite{chen2022self} learns image-specific prototypes by aggregating structure-aware seed regions determined by CAM maps and pixel feature similarities. ~\cite{du2022weakly} constructs class prototypes by aggregating pixel features with top-K CAM scores, enhancing discriminative visual representations by aligning pixels with their class prototypes. Unlike these feature aggregation methods, the proposed approach explicitly learns class representations using multiple class tokens. Transformer attention between class tokens and patch tokens captures multi-level semantic correlations, thereby improving class-specific localization maps.

Some methods refine CAMs using pairwise semantic affinities. AffinityNet~\cite{ahn2018learning} learns pixel affinities from CAM pseudo-labels, enabling CAM propagation via random walk. ~\cite{wang2020weakly} uses segmentation-based pseudo-labels for affinity learning. Other works~\cite{zhang2021complementary,wang2020self} leverage feature affinities from classification networks, while~\cite{gao2021ts} explores multi-task affinities for saliency and segmentation. AFA~\cite{ru2022learning} predicts affinities using transformer attention between patches, guided by segmentation pseudo-labels. ~\cite{lin2023clip} introduces class-aware affinity, applying class-specific masks to transformer attention maps. These approaches enhance CAMs by leveraging semantic affinities for better localization. Although existing methods improve CAM generation mechanisms through image augmentation and specific network architectures, they often overlook the impact of additional class information and contextual information on the network. The proposed method fully exploits these aspects to further enhance WSSS performance.

\subsection{Transformers for Visual Tasks}
Transformers, originally developed for sequential natural language processing (NLP) tasks, have been successfully adapted to visual data, demonstrating remarkable performance across a wide range of computer vision tasks~\cite{wu2025image,wu2025adaptive,wu2025generative,yao2021non}. 
A significant advancement in this domain is the Vision Transformer (ViT)~\cite{dosovitskiy2020image}, a pioneering visual model that leverages the Transformer architecture by processing image patches. 
In a notable study,~\cite{caron2021emerging} trained a self-supervised ViT and discovered that the attention mechanisms between the class token and patch tokens effectively capture the structural layout of scenes. 
Furthermore,~\cite{gao2021ts} enhanced the ViT by integrating a CAM module, enabling class-discriminative localization within the ViT framework. \tred{To address sample efficiency and improve distillation performance, DeiT~\cite{touvron2021training} introduced a data-efficient training strategy for image transformers, demonstrating that carefully designed Transformer training pipelines can achieve competitive results even with limited data.}
\tred{In addition, SegFormer~\cite{xie2021segformer} proposed a simple yet powerful hierarchical Transformer architecture tailored for semantic segmentation, showing that lightweight Transformers with efficient token mixing can achieve strong performance without relying on complex decoder designs.}
\tred{These works further highlight the effectiveness of Transformer-based feature enhancement and hierarchical modeling in dense prediction tasks.}

Recently, researchers have applied ViT to Weakly Supervised Semantic Segmentation (WSSS). 
~\cite{ru2022learning,ru2023token,xu2024mctformer+} generate localization maps through ViT's self-attention mechanisms. 
Specifically, MCTformer+~\cite{xu2024mctformer+} utilizes class-to-patch attention across different class tokens to capture class-specific localization information, while its patch-to-patch attention mechanism effectively learns pairwise affinities to refine localization maps. \tred{Unlike MCTformer, which injects multiple class tokens throughout all Transformer layers, our CPF-CTE introduces class tokens only after the ViT encoder as a post-hoc semantic refinement module, avoiding early-stage class competition and providing cleaner class-conditioned specialization.}
AFA~\cite{ru2022learning} generates reliable affinity labels from pseudo-labels, enforces these affinity labels to supervise the multi-head self-attention mechanism, and ultimately produces robust affinity predictions. 
ViT-PCM~\cite{rossetti2022max} developed a CAM-agnostic, end-to-end solution using the Vision Transformer (ViT) architecture to estimate pixel-level label probabilities, despite the inherent risk of patch misclassification. \bbl{Beyond the visual domain, the Audio-Visual Keyword Transformer (AVKT)~\cite{li2024audio} also employs a Transformer architecture with learnable classification tokens for cross-modal feature aggregation and position-agnostic localization. Although both AVKT and our CPF-CTE utilize learnable class tokens, our design is fundamentally different in that CPF-CTE leverages class tokens as a post-hoc semantic enhancement mechanism specifically tailored for WSSS, enabling refined class activation and improved spatial context fusion within patch representations.}

In this work, we also explore the application of ViT for WSSS and innovatively introduce learnable class tokens to effectively capture class-specific semantic information.


\section{Methodology}

\label{sec:method}

In this section, we describe the overall architecture and key components of our proposed approach. First, in Section~\ref{sec:Overview}, we provide an overview of our framework. Next, in Section~\ref{sec:class_info_enhancement}, we present in detail our novel method of introducing learnable class tokens, which enhances the network by incorporating learnable category-specific information. Finally, in Section~\ref{sec:context_fusion}, we integrate CF-BiLSTM into the network to improve the learning capability by capturing contextual relationships and strengthening information exchange between patches, ultimately boosting the performance of semantic segmentation. Finally, we introce the final prediction in Section~\ref{sec:token}.

\begin{figure*}[t]
\centering
\includegraphics[width=1.0\linewidth]{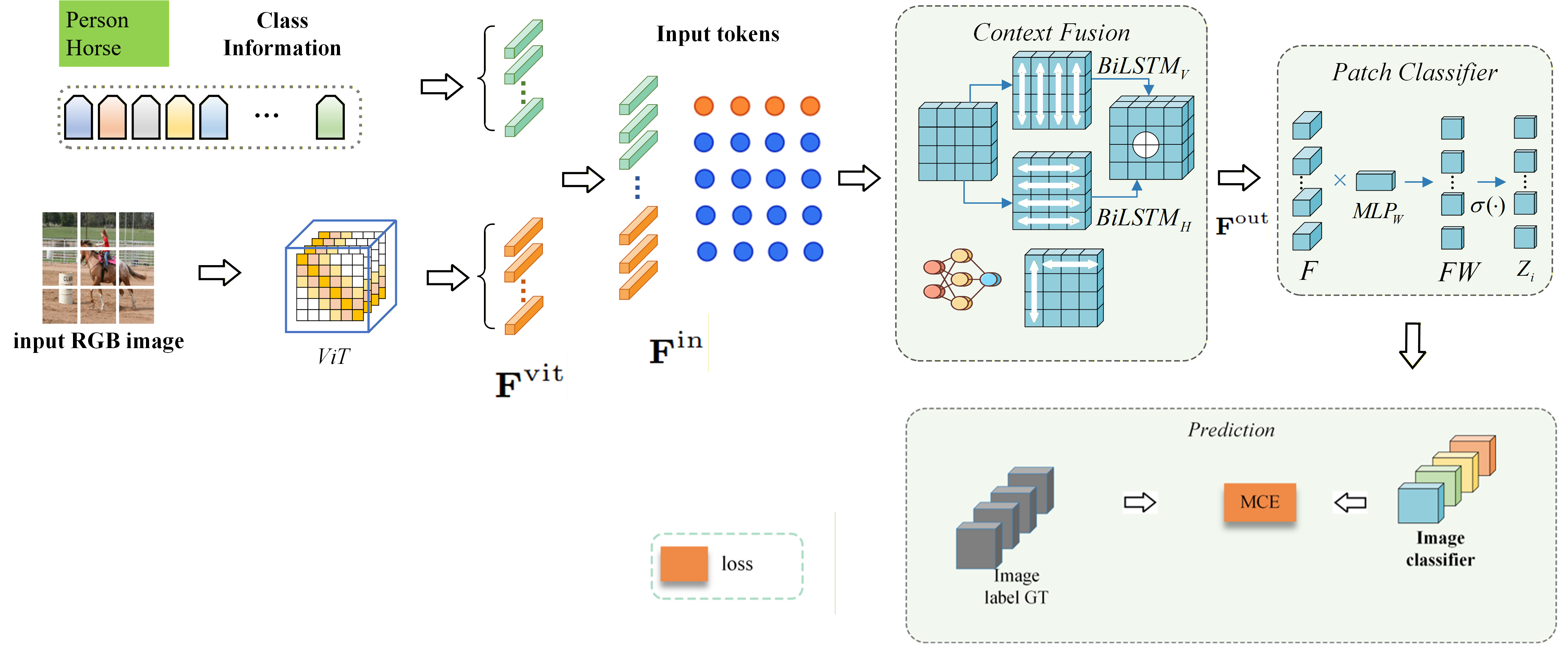}

\caption{
\tred{Overview of the proposed CPF-CTE framework. 
Given an input RGB image, a ViT encoder extracts patch-level features 
$\mathbf{F}^{\text{vit}}$, which are then concatenated to form the input token 
sequence $\mathbf{F}^{\text{in}}$. 
Class information is encoded into learnable class tokens that are processed in 
parallel. 
The CF-BiLSTM module performs contextual fusion through both vertical 
(BiLSTM$_V$) and horizontal (BiLSTM$_H$) bidirectional propagation, producing 
refined representations $\mathbf{F}^{\text{out}}$. 
A patch classifier (MLP$_W$ followed by a sigmoid function) generates 
patch-level activation maps. 
These predictions are further aggregated with the image-level classifier and 
optimized using the multi-class entropy (MCE) loss. }
}
\label{fig:framework}
\vspace{-0.3cm} 
\end{figure*}

\subsection{Overall Framework}
\label{sec:Overview}

\tred{The CPF-CTE framework is composed of three main components: a Vision Transformer (ViT) backbone, a Context Fusion Bidirectional LSTM (CF-BiLSTM) module, and a class-token–enhanced patch classification head. The full pipeline is illustrated in Fig.~\ref{fig:framework}.}

\tred{\noindent\textbf{Patch Tokenization and ViT Encoding.}
The input image is first divided into fixed-size patches, each of which is linearly projected into a token embedding. A learnable class token is added to the sequence, and the entire token set is processed by the ViT encoder. This produces globally contextualized token representations that capture long-range dependencies across the image.}

\tred{\noindent\textbf{Context Refinement via CF-BiLSTM.}
The output tokens from the ViT encoder are then passed into a lightweight Bidirectional LSTM module. This module refines the token features by modeling spatial continuity and exchanging contextual information along the patch sequence. It outputs an updated set of refined patch embeddings, while the class token is also updated through the same interaction.}

\tred{\noindent\textbf{Patch Classification and Pseudo-Label Construction.}
Each refined patch embedding is fed into a lightweight classification head to produce patch-level predictions. These predictions are rearranged back into spatial maps to construct initial pseudo-labels. A CRF post-processing step is applied to sharpen boundaries and remove isolated responses, generating improved pseudo-label masks.}

\tred{\noindent\textbf{Segmentation Model Training.}
The refined pseudo-labels produced by the above pipeline are used as supervisory signals to train a DeepLabv2 segmentation network. The final segmentation outputs are generated entirely by the DeepLabv2 model trained on these pseudo labels.}

\tred{In summary, the CPF-CTE framework processes an image through the following stages: patch tokenization, global feature extraction using ViT, local contextual refinement using CF-BiLSTM, patch-level classification with class token enhancement, CRF refinement, and final segmentation training using DeepLabv2. }

\subsection{Class Information Enhancement}
\label{sec:class_info_enhancement}

To dynamically capture and refine class-specific semantic details in weakly supervised semantic segmentation settings where pixel-level annotations are unavailable, we propose a novel mechanism using learnable class tokens. \tred{Importantly, introducing class tokens at this stage avoids the dominance 
issue commonly observed when multiple class tokens participate in all 
Transformer layers. Instead of influencing early self-attention, our class 
tokens operate on top of the stabilized ViT features, functioning purely as 
class-specific semantic refinement modules.} These tokens act as dynamic, adaptive carriers of discriminative class-specific information, enabling the model to better distinguish between ambiguous regions through explicit class-patch relationship modeling. For a dataset comprising $C$ semantic classes, we initialize a set of learnable tokens denoted as $\mathbf{T} = \{\mathbf{t}_1, \mathbf{t}_2, \dots, \mathbf{t}_C\} \in \mathbb{R}^{C \times H}$, where each token $\mathbf{t}_c \in \mathbb{R}^H$ represents class $c$, and $H$ signifies the dimension of the tokens. These tokens are optimized during training through gradient descent, learning to encode both inter-class distinctions and intra-class commonalities.\tred{Note that these class tokens do not encode textual or linguistic semantics; instead, they learn class-discriminative prototypes through training supervision, enabling semantic refinement without relying on language-based embeddings.}

Unlike conventional class token usage in Vision Transformers (ViT), where a single token aggregates global features for image-level classification, our approach introduces \textbf{multi-class token learning}—a more fine-grained mechanism that allows each category to maintain its own learnable representation. This design is particularly suitable for weakly supervised settings, where class boundaries are uncertain, and class interactions are complex. By learning independent semantic prototypes for each class, the network gains the ability to disentangle overlapping visual concepts and suppress class confusion in mixed or cluttered regions.

Given an input image $\mathbf{I}$, we divide it into $s$ non-overlapping patches and encode each patch with a ViT backbone, obtaining embeddings $\mathbf{P}={\mathbf{p}_1,\dots,\mathbf{p}_s}$, where $\mathbf{p}_i\in\mathbb{R}^e$.
ViT’s self-attention models both local and global dependencies among patches, producing context-enriched representations:
\begin{equation}
\text{Attention}(\mathbf{Q},\mathbf{K},\mathbf{V})
=\text{Softmax}\left(\frac{\mathbf{Q}\mathbf{K}^T}{\sqrt{e}}\right)\mathbf{V}.
\end{equation}
These refined embeddings provide the foundation for subsequent semantic refinement.

\tbl{
However, relying solely on self-attention often leads to category-agnostic feature learning, where the same region may activate for multiple classes due to overlapping visual patterns. To overcome this limitation, CPF-CTE employs learnable class tokens to infuse \textbf{category-specific priors} directly into the patch representations. Intuitively, these tokens act as semantic anchors that guide the network toward discriminative feature learning, bridging the gap between category-level semantics and spatial-level representations.}

Diverging from standard ViT approaches that integrate class tokens at the input stage, we adopt a post-hoc integration strategy. After obtaining the refined patch embeddings $\mathbf{P}^{\text{out}} = \{\mathbf{p}_1^{\text{out}}, ..., \mathbf{p}_s^{\text{out}}\}$ from the final transformer block, each patch embedding $\mathbf{p}_i^{\text{out}}$ is associated with its corresponding class token $\mathbf{t}_c$ through channel-wise concatenation:
\begin{equation}
    \mathbf{f}_i^{\text{in}} = \text{Concat}(\mathbf{p}_i^{\text{out}}, \mathbf{t}_c) \in \mathbb{R}^{e + H},
\end{equation}
where $\mathbf{f}_i^{\text{in}}$ serves as the input to subsequent layers for class-aware feature refinement. \tred{Because class embeddings are concatenated to each patch token and injected at
every LSTM step, class-conditioned gradients do not rely on long recurrent chains,
which avoids gradient diffusion and preserves class-specific semantic distinctions.} \tred{Here, the “dynamic’’ property refers to the learned class-conditioned refinement
behavior rather than explicit visual attention modulation; the effectiveness of this
semantic enhancement is validated through ablation studies rather than attention-map
interpretation.} This post-ViT concatenation ensures that each patch retains its independently learned feature representation while being enriched with class-specific information. The incorporation of class tokens at this stage allows the network to dynamically emphasize relevant class-specific attributes, ultimately leading to more precise segmentation outputs.

This post-hoc fusion strategy offers two major advantages. First, it decouples visual feature learning from semantic conditioning, allowing ViT to focus purely on spatial reasoning while class tokens handle semantic specialization. Second, since the integration occurs after ViT, the class tokens operate on high-level, context-enriched patch embeddings, making their influence more targeted and interpretable. This design contrasts with prior transformer-based WSSS methods, which inject class tokens during early attention computation, often leading to competition among categories during feature aggregation. By contrast, our approach applies semantic enhancement after structural encoding, resulting in a cleaner and more stable category-specific refinement process.

\tred{Although the channel-wise concatenation increases the temporary feature dimensionality,
we immediately apply a learnable linear projection to map it back to the original embedding
size, ensuring dimensional consistency and preventing any mismatch with subsequent layers.}

By explicitly modeling the interactions between classes and patches in later stages, our approach effectively resolves semantic ambiguity and substantially improves segmentation accuracy. This method ensures that even in weakly supervised settings, where precise annotations may be limited, the network can still leverage the class tokens to make more informed predictions. Additionally, the integration of class tokens provides a structured way for the model to encode and utilize class-level information, ultimately leading to more robust feature learning and enhanced segmentation performance.

\tbl{
Furthermore, learnable class tokens can be interpreted as \textbf{semantic prototypes} that evolve during training. Each token gradually learns to represent the central tendency of its corresponding class in the embedding space, promoting intra-class compactness and inter-class separability. This prototype-like behavior improves the model's discriminative ability, especially in scenarios where multiple classes share similar textures or colors. As a result, the class-token mechanism not only strengthens patch-level recognition but also provides an interpretable pathway for visualizing category activations and understanding the decision process of the network.}

\subsection{Contextual Fusion}
\label{sec:context_fusion}
\bbl{The CF-BiLSTM module explicitly restores spatial continuity in ViT feature maps by performing bidirectional contextual fusion along two complementary spatial directions. As illustrated in Fig.~\ref{fig:framework}, BiLSTM$_v$ and BiLSTM$_h$ propagate information vertically and horizontally, enabling message passing between spatially adjacent patches. This design directly mitigates the patch discontinuity issue introduced by ViT tokenization and enhances inter-patch semantic consistency with negligible computational overhead.}

Given the concatenated tokens $ \mathbf{F}^{\text{in}} = [\mathbf{f}_1^{\text{in}}, \dots, \mathbf{f}_s^{\text{in}}] \in \mathbb{R}^{s \times (e+H)} $, the CF-BiLSTM processes them bidirectionally to capture both forward and backward spatial contexts. This bidirectional structure allows the model to integrate spatial dependencies more effectively, reducing inconsistencies in feature representations across patches. By learning long-range dependencies, CF-BiLSTM mitigates fragmented or disjointed feature maps, which are common challenges in patch-based vision models. \tred{Before being fed into the CF-BiLSTM, all patch tokens are arranged in the same
row-major (raster-scan) spatial order used by the ViT patch embedding, ensuring 
consistent adjacency structure and full reproducibility.}

\tred{Following the standard ViT patch embedding order, all patch tokens are arranged in a row-major (raster-scan) sequence before being fed into the CF-BiLSTM. This ensures consistent positional alignment with ViT’s inherent spatial ordering and preserves the adjacency structure encoded during transformer processing.} \tred{Because the forward and backward hidden states are computed by iteratively aggregating neighborhood information, CF-BiLSTM naturally reconstructs missing or weak contextual cues without requiring explicit attention visualization.}

For each token $ \mathbf{f}_i^{\text{in}} $, the forward LSTM ($ \overrightarrow{\text{LSTM}} $) and backward LSTM ($ \overleftarrow{\text{LSTM}} $) compute hidden states by aggregating information from preceding and subsequent patches, respectively:
\begin{equation}
    \overrightarrow{\mathbf{h}}_i = \overrightarrow{\text{LSTM}}(\mathbf{f}_i^{\text{in}}, \overrightarrow{\mathbf{h}}_{i-1}),
\end{equation}
\begin{equation}
    \overleftarrow{\mathbf{h}}_i = \overleftarrow{\text{LSTM}}(\mathbf{f}_i^{\text{in}}, \overleftarrow{\mathbf{h}}_{i+1}),
\end{equation}
where $ \overrightarrow{\mathbf{h}}_i, \overleftarrow{\mathbf{h}}_i \in \mathbb{R}^d $ denote the hidden states in the forward and backward passes, respectively. These hidden states collectively capture rich contextual cues by synthesizing information across adjacent patches. By fusing the outputs of both directional LSTMs, we obtain the final context-enhanced representation for each patch:
\begin{equation}
\mathbf{h}_i = \text{Concat}(\overrightarrow{\mathbf{h}}_i, \overleftarrow{\mathbf{h}}_i) \in \mathbb{R}^{2d}.
\end{equation}

This fusion step ensures that the model effectively retains both past and future spatial dependencies, which enhances the discriminative power of the learned features. By integrating bidirectional contextual information, the model is able to capture long-range dependencies between different patches, thereby enriching the semantic representation of each region in the image. This capability is particularly valuable in vision tasks that require precise spatial understanding, such as semantic segmentation and object detection.

An important advantage of the CF-BiLSTM lies in its \textbf{context reconstruction capability}. When neighboring patches exhibit occlusion or incomplete activation—a common issue in weakly supervised segmentation—bidirectional aggregation allows the model to reconstruct missing contextual evidence by drawing information from both preceding and succeeding spatial regions. This leads to smoother class boundaries and more consistent pseudo masks. Furthermore, the sequential propagation of hidden states implicitly enforces spatial regularization, thereby acting as a soft constraint that reduces over-segmentation artifacts.

Furthermore, the fusion mechanism mitigates potential inconsistencies in feature encoding caused by local ambiguities or occlusions within individual patches. By aggregating information from surrounding patches, the model can reconstruct missing or uncertain details, leading to more stable and coherent feature representations. As a result, the CF-BiLSTM significantly improves the robustness and contextual consistency of patch representations, making it particularly beneficial for complex vision applications. To further refine the extracted information, the fused hidden states undergo a transformation process to ensure compatibility with subsequent layers. The refined token $ \mathbf{f}_i^{\text{out}} $ is obtained by concatenating these states and projecting them back to the original dimension:
\begin{equation}
\mathbf{f}_i^{\text{out}} = \mathbf{W}_c [\overrightarrow{\mathbf{h}}_i; \overleftarrow{\mathbf{h}}_i] + \mathbf{b}_c \in \mathbb{R}^{e+H},
\end{equation}
where $\mathbf{W}_c$ and $\mathbf{b}_c$ are learnable parameters responsible for transforming the concatenated hidden states into a refined token representation. This learnable projection ensures that the bidirectional contextual information is effectively embedded into the final output while maintaining the original feature dimensionality.

In terms of computational complexity, CF-BiLSTM introduces only linear growth with respect to the number of tokens ($\mathcal{O}(s)$), which is considerably more efficient than the quadratic complexity of self-attention ($\mathcal{O}(s^2)$). This makes it an ideal addition to transformer-based segmentation frameworks where efficiency is crucial. Moreover, unlike convolution-based context modules that require fixed receptive fields, CF-BiLSTM dynamically adapts to image structure, enabling the model to flexibly learn context length according to the scene layout.

The resulting output tokens $ \mathbf{F}^{\text{out}} = [\mathbf{f}_1^{\text{out}}, \dots, \mathbf{f}_s^{\text{out}}] $ encode rich contextual relationships, enabling the model to resolve ambiguities in regions with overlapping class semantics. By explicitly modeling spatial dependencies, the CF-BiLSTM facilitates more accurate region classification, particularly in scenarios where class boundaries are difficult to delineate. This bidirectional fusion not only preserves both local and global contextual cues but also enhances the model's ability to generalize across diverse image conditions. Consequently, the CF-BiLSTM serves as a powerful enhancement to the feature extraction pipeline, reinforcing the spatial coherence of patch embeddings and improving overall model performance in vision-based tasks. \tred{Moreover, since CF-BiLSTM operates with linear complexity $\mathcal{O}(s)$ and adds only a 
small number of additional parameters, the module introduces negligible memory and 
runtime overhead compared with the ViT backbone, ensuring scalability to large datasets 
and high-resolution inputs.} \tred{Although BiLSTM involves sequential computation, the overhead is negligible in
practice because it is applied only once on a small set of patch tokens, while the ViT
backbone still dominates the overall runtime.}

Overall, the CF-BiLSTM bridges the gap between global transformer attention and local sequential reasoning, forming a lightweight yet effective contextual fusion mechanism. Its integration within the CPF-CTE framework ensures that spatial coherence, semantic precision, and computational efficiency are jointly optimized, yielding more stable pseudo labels and superior segmentation performance under weak supervision.

\subsection{Final Prediction}
\label{sec:token}

The refined tokens $ \mathbf{F}^{\text{out}} \in \mathbb{R}^{s \times (e+H)} $ are subsequently fed into a lightweight patch classifier designed to generate semantic segmentation predictions with minimal computational overhead. This classifier consists of a single fully connected layer parameterized by the weight matrix $ \mathbf{W} \in \mathbb{R}^{(e+H)\times C} $, followed by a softmax activation function:
\begin{equation}
    \mathbf{Z} = \text{softmax}(\mathbf{F}^{\text{out}} \mathbf{W}),
\end{equation}
where $ \mathbf{Z} \in \mathbb{R}^{s \times C} $ represents the class probability distributions for each individual patch. The softmax operation ensures that the predicted scores for all classes sum to one, enabling direct interpretation as probabilistic confidence scores.

\tbl{
The use of a lightweight classifier at this stage is intentional. Since the ViT backbone and CF-BiLSTM have already captured rich contextual and spatial relationships, a shallow classifier is sufficient to transform refined embeddings into discriminative semantic scores. This design choice avoids unnecessary computational overhead while ensuring that the high-level features produced by CPF-CTE are efficiently converted into pixel-level class evidence.}

To align the patch-level predictions with image-level labels in a weakly supervised learning setting, we employ the Top-$k$ pooling technique~\cite{wu2024top}. This method selects the most confident patches per class, aggregating their scores to form a more reliable image-level prediction. Formally, the image-level confidence score for class $ c $ is computed as:
\begin{equation}
    p_c = \frac{1}{k} \sum_{i=1}^{k} \text{Top-k}(\mathbf{Z}_{j}^{c}), \quad j \in \{1, \ldots, s\},
\end{equation}
where $ p_c $ denotes the aggregated confidence score for class $ c $. The Top-$k$ pooling mechanism effectively mitigates the impact of noisy or misclassified patches, ensuring that the final prediction is primarily influenced by the most reliable evidence present in the image.

\tbl{
This Top-$k$ aggregation plays a critical role under weak supervision. In the absence of pixel-level labels, many patches may correspond to uncertain background regions or ambiguous object boundaries. By selectively averaging only the most confident activations, the model focuses on high-precision cues, thereby reducing noise propagation and false positives. Moreover, the pooling operation implicitly encourages discriminative feature learning, as the model must generate a small number of highly confident activations for each positive class to minimize the loss.}

By leveraging this approach, the model is able to robustly integrate patch-level semantics into a coherent image-level understanding, which is particularly beneficial in scenarios where only weak supervision is available. Weakly supervised segmentation tasks often suffer from noisy annotations and incomplete supervision signals, making it crucial for the model to effectively aggregate discriminative features from reliable patch representations. By incorporating spatial dependencies through the CF-BiLSTM module, our method ensures that contextual information is preserved, reducing fragmentation in feature maps and leading to more holistic segmentation outputs. 

\tbl{
In practice, the combination of CF-BiLSTM and Top-$k$ pooling creates a complementary learning effect. CF-BiLSTM enforces contextual smoothness by enhancing inter-patch consistency, while Top-$k$ pooling selectively amplifies the most confident category evidence. Together, these modules produce pseudo masks that are both contextually coherent and semantically precise, which are ideal for supervising the subsequent fully supervised refinement stage.}

This strategy not only enhances segmentation accuracy but also helps in reducing ambiguity in class assignments, leading to more precise and interpretable results. The incorporation of bidirectional context enables the model to better resolve class inconsistencies that commonly occur in weakly supervised learning settings, where similar structures may be assigned different labels due to lack of strong pixel-level annotations. Furthermore, by refining the learned representations, our approach improves the network's generalization ability, making it more robust to variations in input images.

To supervise the training of our network, we utilize a multi-label classification loss, denoted as $ \mathcal{L}_{MCE} $, which is computed as follows:
\begin{equation}
    \mathcal{L}_{MCE} = -\frac{1}{C}\sum_{c=1}^{C} y_c \log(p_c) + (1-y_c) \log(1-p_c),
\end{equation}
where $ y_c $ is the ground-truth label for class $ c $, representing whether the class is present in the image. The term $ p_c $ corresponds to the predicted confidence score for class $ c $, obtained through the Top-$k$ pooling mechanism. This loss function encourages the network to assign higher confidence scores to the correct classes while minimizing incorrect activations, thereby improving classification reliability and segmentation consistency.

\tbl{
This multi-label formulation aligns naturally with WSSS objectives. Since an image may contain multiple foreground categories, binary cross-entropy across classes avoids mutual exclusivity assumptions and enables the network to learn multi-class co-occurrence patterns. This is particularly beneficial for complex natural scenes (e.g., VOC and COCO), where objects of different categories frequently overlap or appear together.}

\begin{figure}[t]
\centering
\includegraphics[width=1.0\linewidth]{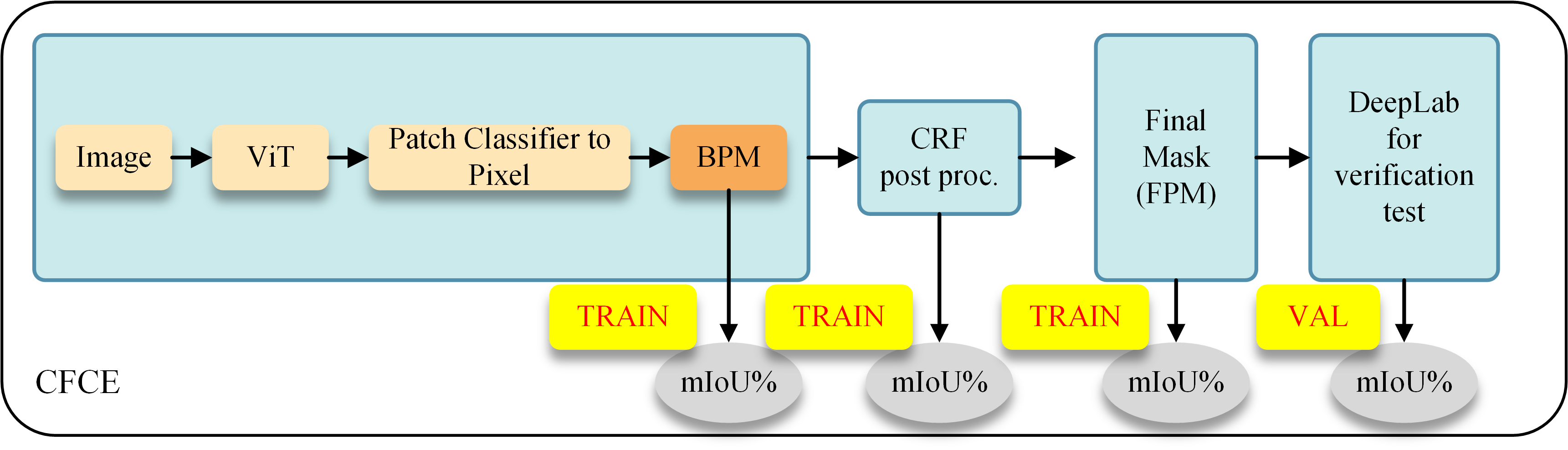}
\caption{
\tred{Pipeline of pseudo-label refinement and final segmentation prediction.  
The patch classifier outputs a baseline pseudo mask (BPM), which is refined by CRF to obtain cleaner pseudo labels.  
These refined pseudo labels are then used to train a DeepLabv2 segmentation network, whose output forms the final prediction.  }
}
\label{fig:finalprediction}
\end{figure}

In the inference phase, the patch predictions $\mathbf{Z}$ are upsampled to the original image resolution using bilinear interpolation, ensuring that the spatial structure of the predictions aligns well with the input image. To generate final segmentation masks, we apply an argmax operation over the upsampled probability maps:
\begin{equation}
    \hat{Y}(i,j) = \arg\max_c \mathbf{Z}(i,j, c),
\end{equation}
where $ \hat{Y}(i,j) $ represents the predicted class label for pixel $(i,j)$. This step converts the soft predictions into discrete semantic labels, forming the final segmentation map.

\tbl{
After generating the baseline pseudo mask (BPM), a Conditional Random Field (CRF)~\cite{kolesnikov2016seed} is applied to refine spatial boundaries and eliminate spurious activations. The CRF acts as a low-level structural prior, promoting local smoothness while preserving sharp object edges, which are often blurred in weakly supervised masks. This refinement step is lightweight yet crucial—it bridges the gap between patch-level confidence maps and dense pixel-wise annotations, significantly improving the usability of pseudo labels for downstream training.}

As illustrated in Fig.~\ref{fig:finalprediction}, the final segmentation prediction process begins with a patch classifier that maps patches to pixel-level predictions, forming the baseline pseudo mask (BPM). To enhance its accuracy, BPM is further refined using Conditional Random Fields (CRF). The refined BPM then serves as supervision for training a fully supervised DeepLab network, ultimately producing the final semantic segmentation output. 

\tbl{
This two-stage prediction–refinement pipeline combines the efficiency of weak supervision with the precision of fully supervised training. In the first stage, CPF-CTE efficiently generates category-aware pseudo masks using only image-level labels. In the second stage, these refined masks supervise a DeepLabv2 model, which learns detailed spatial structures and produces high-resolution segmentation results. This hybrid design not only improves the segmentation quality but also demonstrates that well-structured pseudo labels can serve as strong surrogates for dense supervision.}

The end-to-end design of our framework ensures a balanced integration of local details and global semantics, leading to precise and reliable segmentation results. By effectively incorporating multi-scale contextual information while maintaining computational efficiency, our method offers a practical and scalable solution for weakly supervised semantic segmentation tasks.

\section{Experiment}

\subsection{Experimental Settings}

\noindent\textbf{Dataset.}
Our experiments were conducted using the widely recognized Pascal VOC 2012 dataset~\cite{everingham2010pascal}, a benchmark dataset commonly employed for evaluating semantic segmentation models. The dataset comprises 20 distinct object classes in addition to a background category, covering a diverse set of objects such as animals, vehicles, and household items. Due to its complexity and variety, Pascal VOC 2012 poses significant challenges for segmentation models, requiring them to effectively differentiate between similar-looking objects and background regions. To augment the amount of available training data and enhance model generalization, we adopt the common practice of incorporating additional images from the Semantic Boundaries Dataset (SBD)~\cite{hariharan2011semantic}. The SBD dataset provides extra annotations that complement Pascal VOC 2012, resulting in an expanded training set with 10,582 weakly labeled images, while 1,449 images are designated for validation. 
This extended dataset configuration is widely used in weakly supervised semantic segmentation research and allows for more reliable performance evaluation.
\bbl{Additionally, we further evaluate the generalization capability of our method on the MS COCO 2014 dataset~\cite{lin2014microsoft}, which contains 80 object classes covering a wide range of everyday scenes with diverse object co-occurrence patterns. Following standard WSSS settings, we use the 80K training images with image-level annotations and report results on the 40K validation images. Compared with PASCAL VOC, MS COCO introduces significantly more complex visual layouts, small objects, and cluttered backgrounds, making it a more challenging benchmark for validating the robustness and scalability of weakly supervised segmentation approaches.}

\noindent\textbf{Evaluation Metric.} To quantitatively measure the effectiveness of our proposed segmentation model, we employ the mean Intersection-over-Union (mIoU) as the primary evaluation metric. The mIoU is a widely used standard in semantic segmentation, providing an objective assessment of a model's ability to accurately segment objects and assign correct class labels at the pixel level.

\begin{figure}[t]
\centering
\includegraphics[width=0.6\linewidth]{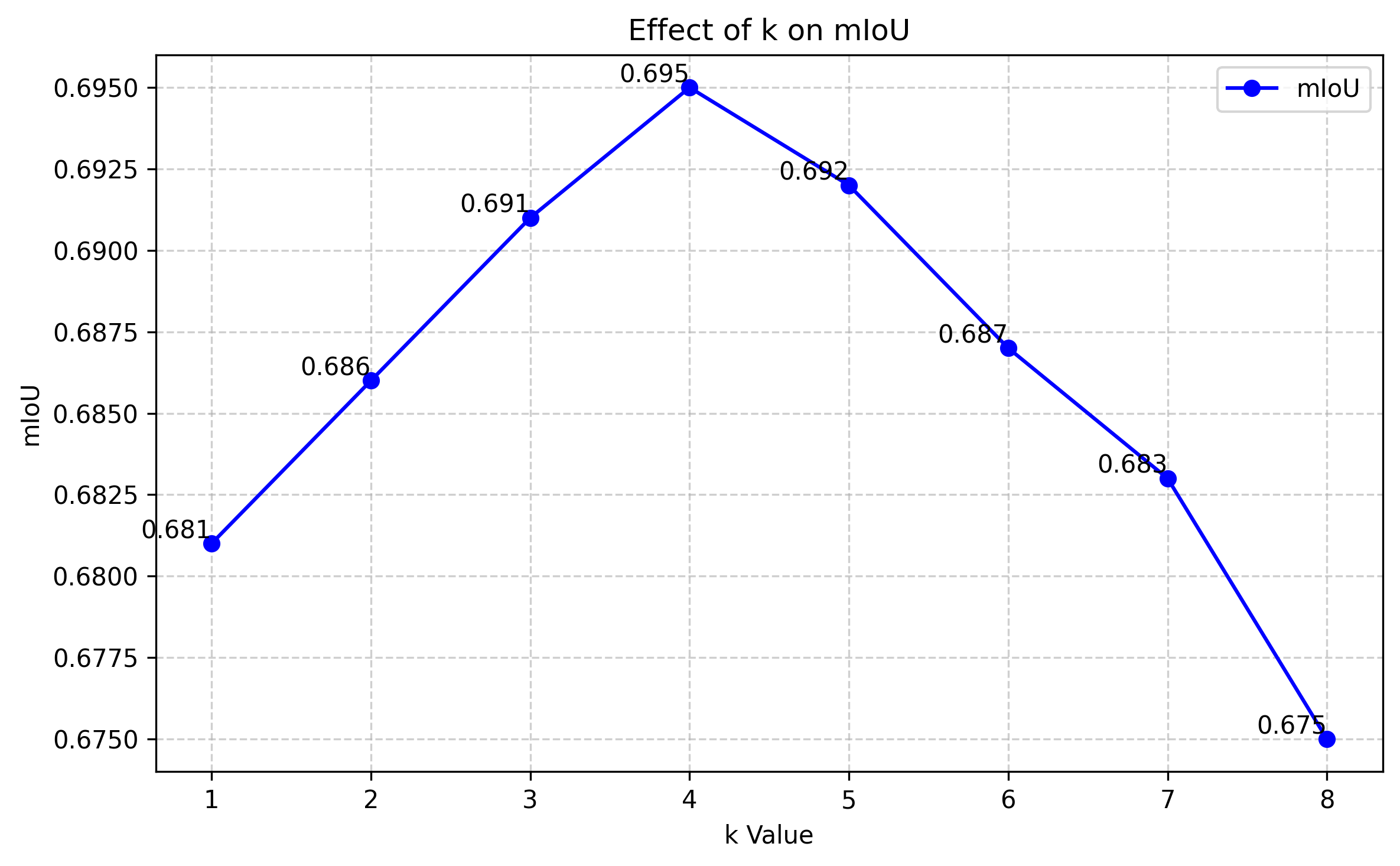}

\caption{The impact of different K values on the final performance of WSSS.}
\label{fig:tkp}

\end{figure}

\subsection{Experimental Implementation Details}
In our experiments, we employ the Vision Transformer (ViT-B/16) model as the encoder, leveraging its powerful representation capabilities for image understanding. To ensure consistency in input dimensions, all images are resized to a resolution of $384 \times 384$ during the training phase, as suggested by \cite{kolesnikov2016seed}. Each image is divided into $24 \times 24$ non-overlapping patches, with each patch having a size of $16 \times 16$ pixels. This patch-based approach allows the model to efficiently capture local and global features within the image.  

The training process is conducted using a batch size of 16 and runs for a maximum of 50 epochs. We utilize two NVIDIA 4090 GPUs to accelerate the training process, ensuring efficient computation and reduced training time. For optimization, we adopt the Adam optimizer, which is known for its effectiveness in handling large-scale datasets and complex models. The learning rate is scheduled in a two-stage manner: an initial learning rate of $10^{-3}$ is applied for the first two epochs to facilitate rapid convergence, followed by a reduced learning rate of $10^{-4}$ for the remaining epochs until the model converges.

\begin{figure}[t]
\centering
\includegraphics[width=0.6\linewidth]{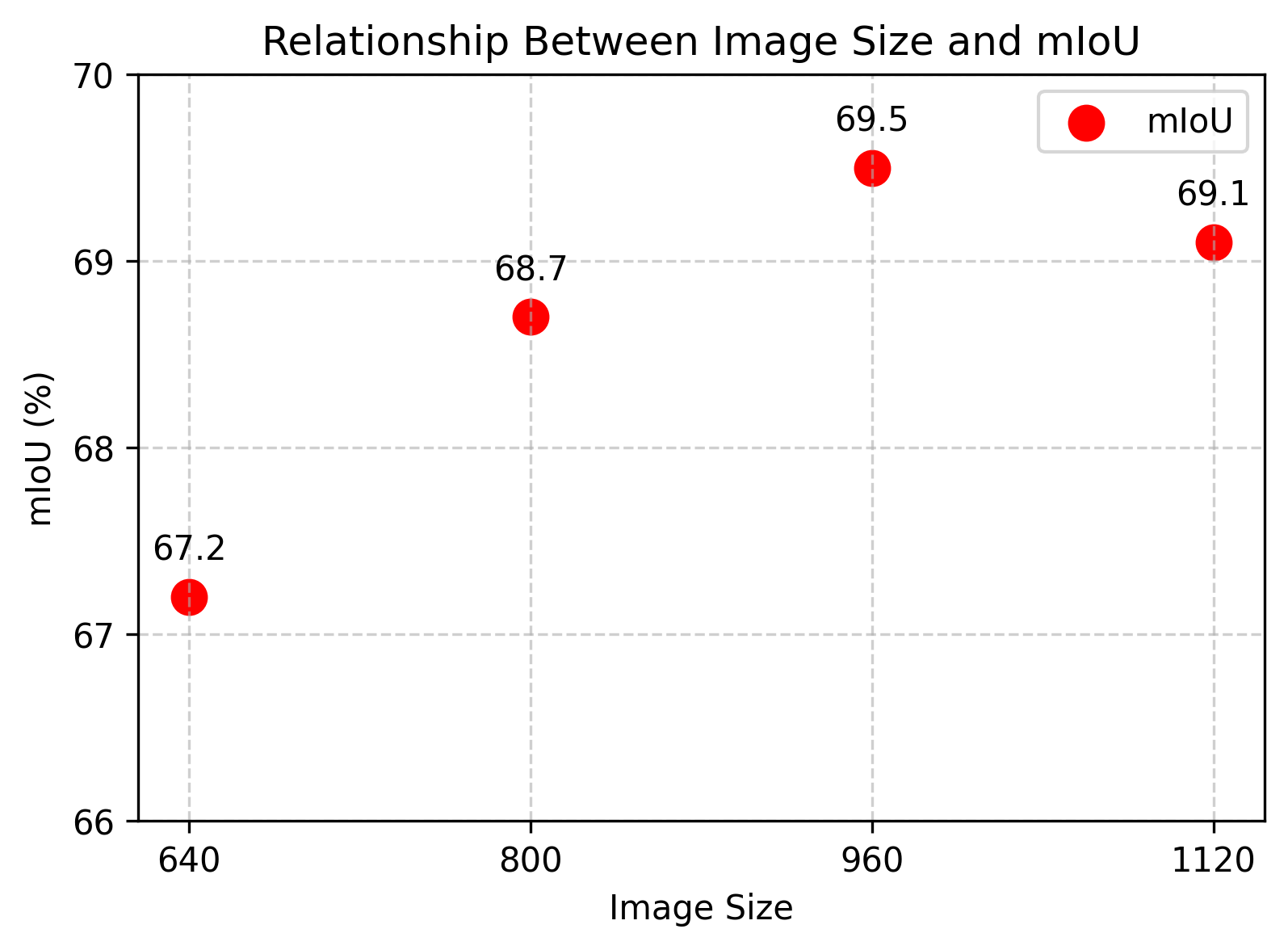}

\caption{The impact of different patch values on the final performance of WSSS.}
\label{fig:patch}

\end{figure}

During the training phase, as illustrated in Fig.~\ref{fig:tkp}, we employ the Top-K pooling strategy with $ k = 4 $ to selectively retain the most informative and discriminative features. This approach ensures that the model focuses on the most confident patch-level predictions while mitigating the influence of noisy or ambiguous regions. By emphasizing high-confidence features, Top-K pooling enhances the robustness of the learned representations, leading to better generalization across diverse and challenging image scenarios. The choice of $ k = 4 $ is based on empirical observations, balancing the trade-off between capturing sufficient contextual information and avoiding overfitting to potentially erroneous activations.

\begin{figure}[t]
\centering
\includegraphics[width=1.0\linewidth]{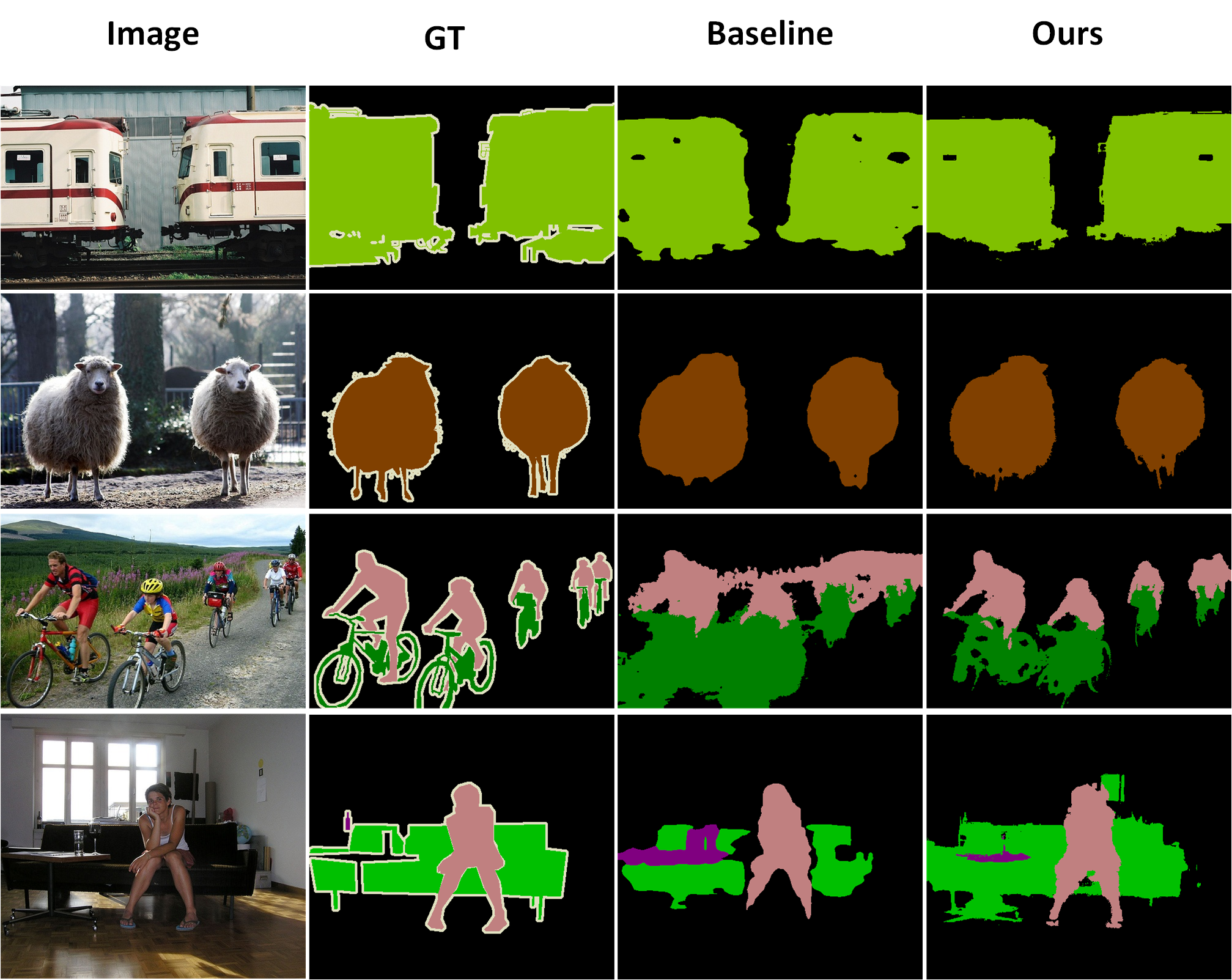}

\caption{Visualization of segmentation results on the val set of PASCAL VOC.}
\label{fig:result}

\end{figure}

For the inference stage, as illustrated in Fig.~\ref{fig:patch},we preprocess the input images by upscaling them to a resolution of $ 960 \times 960 $, which significantly improves the granularity of the segmentation predictions. The use of higher resolution inputs allows the model to capture fine details and intricate structures, thereby refining object boundaries and reducing pixel-level classification errors. This is particularly beneficial in weakly supervised semantic segmentation, where precise boundary delineation is crucial for enhancing segmentation accuracy. Additionally, the high-resolution inference strategy ensures that small or thin objects, which might be overlooked at lower resolutions, are more accurately segmented. 

Overall, the combination of Top-K pooling during training and high-resolution inference constitutes an effective strategy for improving both model robustness and segmentation quality. By focusing on the most confident features and leveraging detailed high-resolution predictions, our method achieves enhanced performance in challenging segmentation tasks.

In the semantic segmentation stage, we utilize the DeepLab V2 framework \cite{chen2018encoder} to train the model. The training is performed using dense pixel pseudo-labels generated in the previous stage, which serve as a supervisory signal for the segmentation task. This approach ensures that the model learns to accurately classify each pixel in the image. Finally, to further refine the segmentation results and improve boundary precision, we apply CRF\cite{kolesnikov2016seed}. The CRF post-processing step helps to smooth the segmentation outputs and align them more closely with the object boundaries, resulting in higher-quality segmentation masks.

\subsection{Comparison with State-of-the-Arts}

\noindent\textbf{Comparison of Pseudo Labels.}
As shown in Table~\ref{tab:vocbpm}, our proposed CPF-CTE achieves the highest pseudo label quality on the PASCAL VOC 2012 dataset, surpassing all existing state-of-the-art (SOTA) approaches. Specifically, CPF-CTE attains a \textbf{mean IoU (mIoU) of 70.8}, outperforming previous Transformer-based methods such as AFA~\cite{ru2022learning} (66.0), PGSD~\cite{hao2024prompt} (68.7), and CGM~\cite{zhang2024enhanced} (68.1). This performance gain is primarily attributed to the synergistic effect of the learnable class token and the context fusion mechanism, which jointly enhance semantic discrimination and contextual reasoning. 

Compared with Convolutional Neural Network (CNN)–based frameworks such as SEAM~\cite{wang2020self}, CDA~\cite{su2021context}, and FPR~\cite{chen2023fpr}, our ViT-B/16 backbone provides stronger global representation and enables more reliable pseudo-label generation under weak supervision. These results validate the robustness and scalability of CPF-CTE in producing high-quality pseudo masks, setting a new benchmark for weakly supervised semantic segmentation.

\begin{figure}[t]
    \centering
    \includegraphics[width=0.8\linewidth]{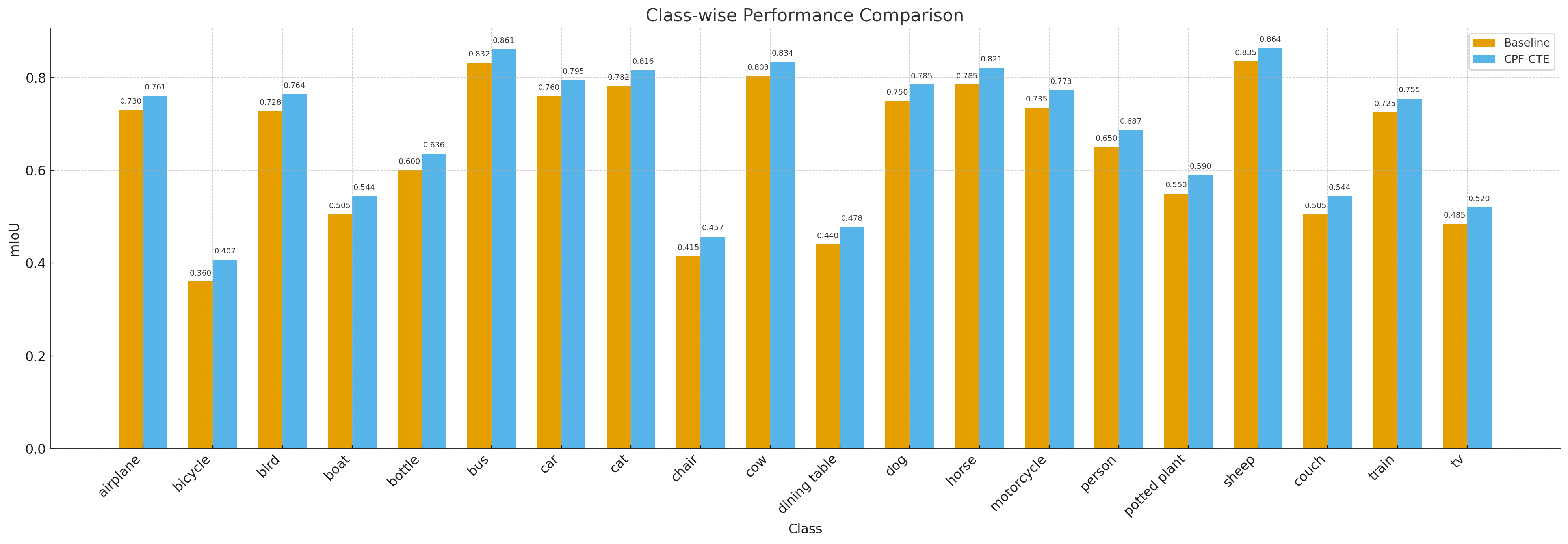}
    \caption{\bbl{Class-wise segmentation performance comparison on the validation set. The figure reports per-class mIoU scores of our CPF-CTE and the baseline model across 20 semantic categories, clearly illustrating the consistent improvements achieved by our method. }}
    \label{fig:classwise}
\end{figure}

\begin{table}[ht]
\centering
\small 
\caption{Pseudo Label Performance Comparison (mIoU) on Pascal VOC 2012 train set. }
\vspace{0.5cm}
\setlength\tabcolsep{12pt} 
\label{tab:vocbpm}
\begin{tabular}{l|l|l|cc}
\toprule
Method & Pub. & Backbone & mIoU (\%) \\
\midrule
SEAM~\cite{wang2020self} & CVPR20 & V1-RN38 & 63.6 \\
CDA~\cite{su2021context} & ICCV21 & V1-RN38 & 66.4 \\
AdvCAM~\cite{lee2022anti} & T-PAMI22 & V2-RN101 & 69.9 \\
SIPE~\cite{chen2022self} & CVPR22 & ResNet50 & 64.4 \\
AFA~\cite{ru2022learning} & CVPR22 & ViT-B/16 & 66.0 \\
FPR~\cite{chen2023fpr} & ICCV23 & ResNet38 & 68.5 \\
PGSD~\cite{hao2024prompt} & TCSVT24 & ViT-B/16 & 68.7 \\
CGM~\cite{zhang2024enhanced}& PR24 & ViT-B/16 & 68.1 \\
\bbl{CPF-CTE w/o CRF} & Ours & ViT-B/16 & 67.1\\
CPF-CTE & Ours & ViT-B/16 & 70.8 \\
\bottomrule
\end{tabular}
\end{table}

\begin{table}[ht]
\centering
\small 
\caption{Semantic Segmentation Performance Comparison (mIoU) on Pascal VOC 2012 val set. }
\vspace{0.5cm}
\setlength\tabcolsep{12pt} 
\label{tab:vocseg}

\begin{tabular}{@{}lllll@{}}

\toprule
Model & Pub. & Backbone & mIoU (\%)\\
\midrule

RRM~\cite{zhang2020reliability} & AAAI20 & ResNet50 & 66.3 \\
BES~\cite{chen2020weakly} & ECCV20 & ResNet50 & 65.7 \\
AFA~\cite{ru2022learning} & CVPR22 & ViT-B/16 & 63.8 \\
SIPE~\cite{chen2022self} & CVPR22 & ResNet50 & 58.6 \\
TSCD~\cite{Xu_Wang_Sun_Xu_Meng_Zhang_2023} & AAAI23 & MiT-B1 & 67.3 \\
PGSD~\cite{hao2024prompt} & TCSVT24 & ViT-B/16 & 68.7 \\
CGM~\cite{zhang2024enhanced}& PR24 & ViT-B/16 & 67.8 \\
CPF-CTE & Ours & ViT-B/16 & 69.5 \\
\bottomrule
\end{tabular}
\end{table}

\begin{table}[t] 
\centering
\small 
\caption{Semantic Segmentation Performance Comparison (mIoU) on MS COCO 2014 val set.}
\label{tab:cocoseg}
\setlength\tabcolsep{12pt} 
\begin{tabular}{@{}llll@{}}
\toprule
Model & Pub. & Backbone & mIoU (\%) \\
\midrule
AFA~\cite{ru2022learning} & CVPR22 & ViT-B/16 & 38.9 \\
SIPE~\cite{chen2022self} & CVPR22 & Resnet38 & 43.6\\
SAS~\cite{kim2023semantic} & AAAI23 & ViT-B/16 & 44.5 \\
FPR~\cite{chen2023fpr} & ICCV23 & ResNet38 & 43.9 \\
ToCo~\cite{ru2023token} & CVPR23 & ViT-B & 42.3 \\
TSCD~\cite{Xu_Wang_Sun_Xu_Meng_Zhang_2023} & AAAI23 & MiT-B1 & 40.1 \\
\bbl{PGSD}~\cite{hao2024prompt} & TCSVT24 & ViT-B/16 & 43.5 \\
\bbl{CGM}~\cite{zhang2024enhanced}& PR24 & ViT-B/16 & 40.1 \\
\bbl{CPF-CTE (pseudo-label)} & \bbl{Ours} & ViT-B/16 & \bbl{41.3} \\
CPF-CTE & Ours & ViT-B/16 & 45.4 \\
\bottomrule
\end{tabular}

\end{table}

\vspace{0.3cm}
\noindent\textbf{Improvements in Segmentation Results.}
To further evaluate the effectiveness of the generated pseudo labels, we train DeepLab V2 using the masks produced by CPF-CTE and compare the segmentation results with previous methods on the PASCAL VOC validation set (Table~\ref{tab:vocseg}). Our method achieves an mIoU of \textbf{69.5}, clearly outperforming all competitors, including the most recent PGSD~\cite{hao2024prompt} (68.7) and CGM~\cite{zhang2024enhanced} (67.8). This improvement demonstrates that CPF-CTE not only enhances pseudo-label quality but also leads to stronger downstream segmentation accuracy.

Notably, CPF-CTE surpasses both CNN- and Transformer-based baselines. For example, compared to AFA~\cite{ru2022learning} (63.8, ViT-B/16), our model yields a substantial margin of +5.7 and +7.8 mIoU, respectively. This advantage arises from our class token design, which explicitly models class-level semantics, and the context fusion strategy, which improves intra-image relational learning. Qualitative visualizations in Fig.~\ref{fig:result} further illustrate that CPF-CTE produces cleaner object boundaries and fewer false activations, particularly in cluttered or occluded regions. \bbl{To further analyze the enhanced discrimination capability, we present the per-class mIoU comparison with a strong baseline model in Fig.~\ref{fig:classwise}. CPF-CTE consistently improves upon the baseline across most categories.} We further evaluate CPF-CTE on the challenging MS COCO 2014 dataset to assess its generalization ability in large-scale and complex scenarios. As shown in Table~\ref{tab:cocoseg}, \bbl{our method achieves an mIoU of \textbf{45.4}, outperforming recent methods such as PGSD~\cite{hao2024prompt} (43.5) and CGM~\cite{zhang2024enhanced} (40.1).} \bbl{We also report the pseudo-label performance in Table~\ref{tab:cocoseg}. The model trained directly on the pseudo-labels achieved 41.3 mIoU prior to the full supervision training. The significant gap between this intermediate baseline and our final result of 45.4 mIoU clearly demonstrates the effectiveness of the two-stage training design in boosting WSSS performance.} This consistent improvement across datasets demonstrates that CPF-CTE generalizes well to diverse object categories and dense multi-object scenes. The performance gain on COCO confirms that our class-token-driven context modeling effectively scales beyond PASCAL VOC and remains robust under more complex image distributions.

\begin{table}[ht]
\centering
\small 
\caption{Ablation Study on the Impact of Class Additional Token and Context Enhancement on Semantic Segmentation Performance}
\vspace{0.5cm}
\label{tab:ablation}
\begin{adjustbox}{width=1.0\linewidth}
\begin{tabular}{lcccc}
\toprule
\textbf{Original Framework} & \textbf{Class Additional Token} & \textbf{Context Enhancement} & \textbf{mIoU (\%)} \\
\midrule
\checkmark & & & 65.2 \\
\checkmark & \checkmark & & 67.1 \\
\checkmark & & \checkmark & 67.8 \\
\checkmark & \checkmark & \checkmark & 69.5 \\
\bottomrule
\end{tabular}
\end{adjustbox}
\end{table}

\subsection{Ablation Studies}
To thoroughly evaluate the impact of our proposed components, we conduct a series of ablation studies, as summarized in Tab.~\ref{tab:ablation}. The baseline framework, which serves as our starting point, achieves a mean Intersection-over-Union (mIoU) of 65.2\% on the validation set. This result provides a reference point for assessing the contributions of the individual components we introduce. \tred{CF-BiLSTM is designed as a complementary post-hoc refinement module rather than a replacement for transformer self-attention; therefore, our ablation focuses on validating its complementary effect rather than conducting direct attention-block substitution experiments.}
 First, we investigate the effect of incorporating the class additional token into the framework. This token is designed to capture class-specific information, enabling the model to better distinguish between different object categories. When the class additional token is added, the mIoU increases by 1.9\%, reaching 67.1\%. This improvement underscores the importance of explicitly modeling class-related features, which helps the network focus on discriminative regions within the image. Next, we examine the impact of context enhancement, which aims to improve the model's ability to capture intra-image contextual relationships. By integrating this component, the framework achieves an mIoU of 67.8\%, representing a 2.6\% improvement over the baseline. This result highlights the significance of understanding spatial dependencies and contextual cues within the image, which are critical for accurate semantic segmentation, especially in complex scenes with overlapping objects or ambiguous boundaries. Finally, we combine both the class additional token and context enhancement into a unified framework. This integrated approach yields the highest performance, achieving an mIoU of 69.5\%. The 4.3\% improvement over the baseline demonstrates the synergistic effect of combining class-specific information with enhanced contextual modeling. Together, these components enable the model to not only identify object categories more accurately but also refine the boundaries and spatial relationships between objects. These results highlight the effectiveness of our approach in enhancing semantic segmentation performance.

\tred{In addition to evaluating the effects of the CF-BiLSTM and class-token enhancement modules, we further analyze different pooling strategies used for patch-level aggregation before generating pseudo labels. As shown in Tab.~\ref{tab:diff-pooling}, we compare three commonly used pooling mechanisms—global average pooling, max pooling, and Top-K pooling—while keeping all other components unchanged. Global average pooling yields an mIoU of 67.7\%, indicating that uniformly aggregating all patch responses tends to dilute discriminative cues, especially in weakly supervised settings where foreground activation is sparse. Max pooling performs better (68.1\%) by preserving the strongest responses, but it is also sensitive to noise and may overemphasize isolated activations. In contrast, Top-K pooling achieves the best performance (69.5\%), demonstrating its ability to balance robustness and selectivity by aggregating only the most confident patch activations while suppressing background noise. This result justifies our choice of Top-K pooling in the final framework and further confirms its benefit in improving pseudo-label quality under weak supervision.}

\begin{table}[ht]
\centering
\caption{\tbl{Ablation studies on different pooling strategies and keeping other components consistent for final semantic segmentation performance on Pascal VOC 2012 val.}}
\label{tab:diff-pooling}

\begin{tabular}{ccccccc}
\toprule
Average pooling & Max-Pooling & Top-K & mIoU(\%)\\
\midrule
\checkmark & &  & 67.7\% \\
& \checkmark &  & 68.1\% \\
&  &  \checkmark & 69.5\% \\
\bottomrule
\end{tabular}

\end{table}

\section{Conclusion}
In this work, we propose a CPF-CTE approach for WSSS. Unlike previous frameworks that rely on single image inputs, we introduce learnable class tokens to effectively represent class-specific information. These tokens are dynamically optimized during training, enabling the model to capture discriminative features for each class without requiring pixel-level annotations. Additionally, we enhance context interaction between patches through a CF-BiLSTM module, which leverages bidirectional dependencies to model long-range spatial relationships within the image. This module not only improves the network's ability to capture intra-image contextual relationships but also addresses the limitations of traditional methods that struggle with complex object boundaries and occlusions. By integrating these two components into a robust baseline, we achieve SOTA results in WSSS using only image-level labels. \tred{Despite its effectiveness, the proposed CPF-CTE framework still relies on patch-level processing, which may limit its scalability for very high-resolution images and densely packed objects. Moreover, the bidirectional LSTM introduces sequential computation that could increase inference latency compared with fully parallel transformer designs. Future work will explore more efficient context modeling mechanisms to further improve scalability.}

\section*{Acknowledgement}
Not applicable.

\section*{Funding Statement}
The authors received no specific funding for this study.

\section*{Author Contributions}

\textbf{Yiyang Fu}: Conceptualization, Methodology, Software, Data curation, Experiments, Visualization, Writing -- original draft. 

\textbf{Hui Li}: Supervision, Methodology, Validation, Writing -- review \& editing, Project administration. 

\textbf{Wangyu Wu}: Conceptualization, Supervision, Formal analysis, Resources, Writing -- review \& editing.

All authors reviewed the results and approved the final version of the manuscript.

\section*{Availability of Data and Materials}
The datasets used in this study, PASCAL VOC 2012, MS COCO 2014, and the Semantic Boundaries Dataset (SBD), are publicly available at 
\url{http://host.robots.ox.ac.uk/pascal/VOC/}, 
\url{https://cocodataset.org/}, 
and 
\url{https://www2.eecs.berkeley.edu/Research/Projects/CS/vision/grouping/semantic_contours/}

\section*{Ethics Approval}
This article does not involve human participants or animals and therefore does not require ethical approval.

\section*{Conflicts of Interest}
The authors declare no conflict of interest.

\bibliographystyle{IEEEtran}
\bibliography{tsp}

\end{document}